\documentclass[review,3p]{elsarticle}
\pdfoutput=1
\usepackage{hyperref}
\usepackage{blindtext}
\usepackage[inline]{enumitem}
\usepackage{lineno,hyperref}
\usepackage[utf8]{inputenc} % allow utf-8 input
\usepackage[T1]{fontenc}    % use 8-bit T1 fonts
\usepackage{float}
\usepackage{url}            % simple URL typesetting
\usepackage{booktabs}       % professional-quality tables
\usepackage{amsfonts}       % blackboard math symbols
\usepackage{nicefrac}       % compact symbols for 1/2, etc.
\usepackage{microtype}      % microtypography
\usepackage{lipsum}
\usepackage{graphicx}
\usepackage[lofdepth,lotdepth]{subfig}
\usepackage{tabu}
\usepackage{subfig}
\usepackage[table]{xcolor}
\usepackage{amsmath}
\usepackage{multirow}
\modulolinenumbers[5]

\journal{Journal of }

\begin{document}

\begin{frontmatter}

\title{Phraseformer: Multimodal Key-phrase Extraction using Transformer and Graph Embedding}

		
\author[mymainaddress]{N. Nikzad\textendash Khasmakhi }
\ead{n.nikzad@tabrizu.ac.ir}
\author[mymainaddress]{Mohammad\textendash Reza Feizi\textendash Derakhshi \corref{mycorrespondingauthor}}
\cortext[mycorrespondingauthor]{Corresponding author}
\ead{mfeizi@tabrizu.ac.ir}
\author[mymainaddress]{Meysam Asgari-Chenaghlu}
\ead{m.asgari@tabrizu.ac.ir}
\author[UT]{M. A. Balafar}
\ead{balafarila@tabrizu.ac.ir}
\author[mymainaddress]{Ali-Reza Feizi-Derakhshi}
\ead{derakhshi96@ms.tabrizu.ac.ir}
\author[mymainaddress,mysecondaddress]{Taymaz Rahkar-Farshi}
\ead{taymazfarshi@ayvansaray.edu.tr}
\author[mymainaddress]{Majid Ramezani}
\ead{m_ramezani@tabrizu.ac.ir}
\author[mymainaddress,myfifthaddress]{Zoleikha Jahanbakhsh-Nagadeh}
\ead{zoleikha.jahanbakhsh@srbiau.ac.ir}
\author[mymainaddress,mythirdaddress]{Elnaz Zafarani-Moattar}
\ead{e.zafarani@iaut.ac.ir}
\author[mymainaddress,myfourthaddress]{Mehrdad Ranjbar-Khadivi}
\ead{mehrdad.khadivi@iaushab.ac.ir}


\address[mymainaddress]{Computerized Intelligence Systems Laboratory, Department of Computer Engineering, University of Tabriz, Tabriz, Iran}
\address[UT]{Department of Computer Engineering, University of Tabriz, Tabriz, Iran}
\address[mysecondaddress]{Department of Software Engineering, Ayvansaray University, Istanbul, Turkey}
\address[myfifthaddress]{Department of Computer Engineering, Naghadeh Branch, Islamic Azad University, Naghadeh, Iran}
\address[mythirdaddress]{Department of Computer Engineering, Tabriz Branch, Islamic Azad University, Tabriz, Iran}
\address[myfourthaddress]{Department of Computer Engineering, Shabestar Branch, Islamic Azad University, Shabestar, Iran.}

\begin{abstract}

\textbf{Background}
Keyword extraction is a popular research topic in the field of natural language processing.  Keywords are terms that describe the most relevant information in a document. The main problem that researchers are facing is how to efficiently and accurately extract the core keywords from a document.  However, previous keyword extraction approaches have utilized the text and graph features, there is the lack of models that can properly learn and combine these features in a best way.

\textbf{Methods}
In this paper, we develop a multimodal Key-phrase extraction approach, namely \textit{Phraseformer}, using transformer and graph embedding techniques. In Phraseformer, each keyword candidate is presented by a vector which is the concatenation of the %representations learned from text and co-occurrence network. 
text and structure learning representations. Phraseformer takes the advantages of recent researches such as BERT and ExEm to preserve both representations. Also, the Phraseformer treats the key-phrase extraction task as a sequence labeling problem solved using classification task.

\textbf{Results}
We analyze the performance of Phraseformer on three datasets including Inspec, SemEval2010 and SemEval 2017 by F1-score. Also, we investigate the performance of different classifiers on Phraseformer method over Inspec dataset. Experimental results demonstrate the effectiveness of Phraseformer method over the three datasets used. Additionally, the Random Forest classifier gain the highest F1-score among all classifiers.
 
\textbf{Conclusions}
Due to the fact that the combination of BERT and ExEm is more meaningful and can better represent the semantic of words. Hence, Phraseformer significantly outperforms single-modality methods.
 
 %In particular, Phraseformer shows better performance than the other compared methods over the three datasets used.

%In the next step, a classifier is built on the concatenation of features. For the classification task, we test four different classifiers. Finally, to verify the effectiveness of BERTERS, we analyze its performance on multi-label classification, recommendation, and visualization tasks. 
\end{abstract}


\end{frontmatter}

%\linenumbers

\section{Introduction}\label{sec:introduction}
There is a phenomenal growth in the quantity of text documents available to users on the different social medias. Hence, it is vital for efficient and effective ways to retrieve and summarize all these documents \cite{vega2019multi}. Keyword or key-phrase extraction is a solution that identifies a set of the terms that conclude the main idea of a document \cite{berry2010text}. This process helps readers to quickly perceive the content of a document. Key-phrase extraction methods can be effectively used by many text  mining  applications such as indexing, visualization, summarization, topic detection  and  tracking, clustering and classification \cite{lahiri2018keywords, zhang2008automatic}.

A number of studies have been conducted to tackle the key-phrase extraction issue. Some of these approaches are textual models that focus only the content of the document to obtain keywords. A group of one have used lexical and syntactic analyses. Other approaches of this class take the advantages of numerical statistics such as term frequency (TF) or term frequency-inverse document frequency (TF-IDF) \cite{wang2019detecting}.  On the other hand, graph-based approaches create a various group of methods by constructing a graph of words. In this class, the most central nodes illustrates keywords \cite{vega2019multi}. Also, we can find hybrid models  that uses the combination of textual and graph-based methods to select keywords. In any case, the topic of which combination method is the most suitable and powerful for catching key-phrases extraction is open for discussion. 

In this paper, we aim to propose a combined method of a graph-based model and a textual model. Also, we view the key-phrases extraction task as classification and sequence tagging problems. After pre-processing part that removes the stop words and punctuation mark, we construct an undirected and unweighted co-occurrence graph for all documents in the corpus. Then, we use three graph embedding techniques consisting ExEm, Node2vec and DeepWalk to learn structure representations of words. On the other hand, each single document is injected to BERT Transformer to get the word embedding. In the next step, we concatenate two embeddings obtained from the graph and the content of the document to create a single representation for each word. After that, we formulate the key-phrase extraction as sequence labeling task that assigns the BIO tagging to each word in a document. Also, our proposed method treats the sequence labeling as a classification task.  The output of the classification is a BIO tag where a word is at the beginning (B) of a key-phrase, inside (I) of a key-phrase and outside of a key-phrase (O). The major contributions of this paper can be summarized as follows:

\begin{itemize}
  \item To the best of our knowledge, this work is the first to apply a multimodal approach to extract key-phrases using Transformer and graph embedding techniques.
  \item Also, to the extent of our knowledge, this is the primary prospective study that directly uses the graph embedding techniques to convert the words in the co-occurrence graph into low-dimensional vectors and employs these vectors to find key-phrases.
 \item We consider the problem of key-phrase extraction from a document as a sequence labeling task. Moreover, we observe the sequence labeling in the form a classification task using the BIO encoding scheme.
\end{itemize}

The rest of the paper is structured as follows: Section \ref{related_work} reviews the related works.  Section \ref{Proposed_Method} presents our proposed method and explains it in detail. 
 The descriptions of our experiments are presented in Section \ref{experiments}.  Section \ref{evaluation_re} provides the experimental results. Finally, Section \ref{sec:conclusion} concludes the paper.

\section{Related Work}\label{related_work}
The exiting techniques used for keyword extraction can fall into three groups: textual, graph-based and hybrid models. Textual approaches generate keywords directly from the original text by applying natural language processing techniques. While, graph-based methods convert the document into a co-occurrence graph where nodes represent words and edges show the relationship between two words in a context window. On the other hand,  hybrid models take the advantage of both text and graph representations of a document to detect keywords. In the following paragraphs, we will investigate these three categories in more detail. In general, the main contributions of our research are summarized as below.

In textual model the aim is to generate keywords directly from the original text \cite{wang2019detecting}. The simplest model in this category uses TF-IDF technique to extract keywords.  After that, researches have focused on machine learning approaches to train a classifier to capture keywords. With the advent of deep learning approaches such Convolutional Neural Networks (CNNs), Recurrent neural network (RNN) architectures such as Long short-term memory (LSTM), and today's Transformers have been popular solutions to this task. There are a number of textual approaches including KEA \cite{witten2005kea}, KP-Miner \cite{el2009kp}, WINGNUS \cite{nguyen2010wingnus}, RAKE \cite{rose2010automatic}, YAKE \cite{campos2020yake},  TNT-KID \cite{martinc2020tnt}, \cite{basaldella2018bidirectional, alzaidy2019bi, tang2019progress, wang2019using, kim2020validation}.

The main idea of graph-based methods is to construct a co-occurrence graph form documents.  The co-occurrence network shows the interactions of words in a corpus.  In this graph, words represent nodes and there is an edge between two words if these words co-occur within a window. 
After constructing the co-occurrence graph, some centrality measures such as degree, closeness, betweenness and eigenvector are applied on it to find keyphrases. In these methods, the keywords are identified by the most central nodes. A number of methods include
TextRank \cite{mihalcea2004textrank}, CollabRank \cite{wan2008collabrank}, DegExt\cite{litvak2011degext}, NE-Rank \cite{bellaachia2012ne}, TopicRank \cite{bougouin-etal-2013-topicrank}, Positionrank \cite{florescu2017positionrank},  M-GCKE \cite{wang2019detecting}, \cite{boudin2013comparison, abilhoa2014keyword, tixier2016graph, el2017graph, boudin2018unsupervised, vega2019multi} that use the graph theory to select keywords. %These methods ignore the semantic representations of nodes in a graph.

The hybrid models attempt to join two previous mentioned categories. These models calculate scores for words from both co-occurrence graph and the document content. Different approaches use various ways to combine these scores. Authors in studies \cite{danesh2015sgrank, zhang2017mike, mahata2018key2vec, mahata2018theme} have proposed hybrid approaches.

Studies \cite{siddiqi2015keyword, merrouni2019automatic, unlu2019survey, papagiannopoulou2020review, firoozeh2020keyword} have conducted reviews of key-phrase and keyword extraction techniques. In our study, by combining the graph-based and text models, and using sequence tagging and classification, we attempt to develop an effective keyword extraction approach that can eliminate the drawbacks of previous studies.

\section{Proposed Method}\label{Proposed_Method}
\begin{figure}
	\centering
	\includegraphics[width=0.8\textwidth]{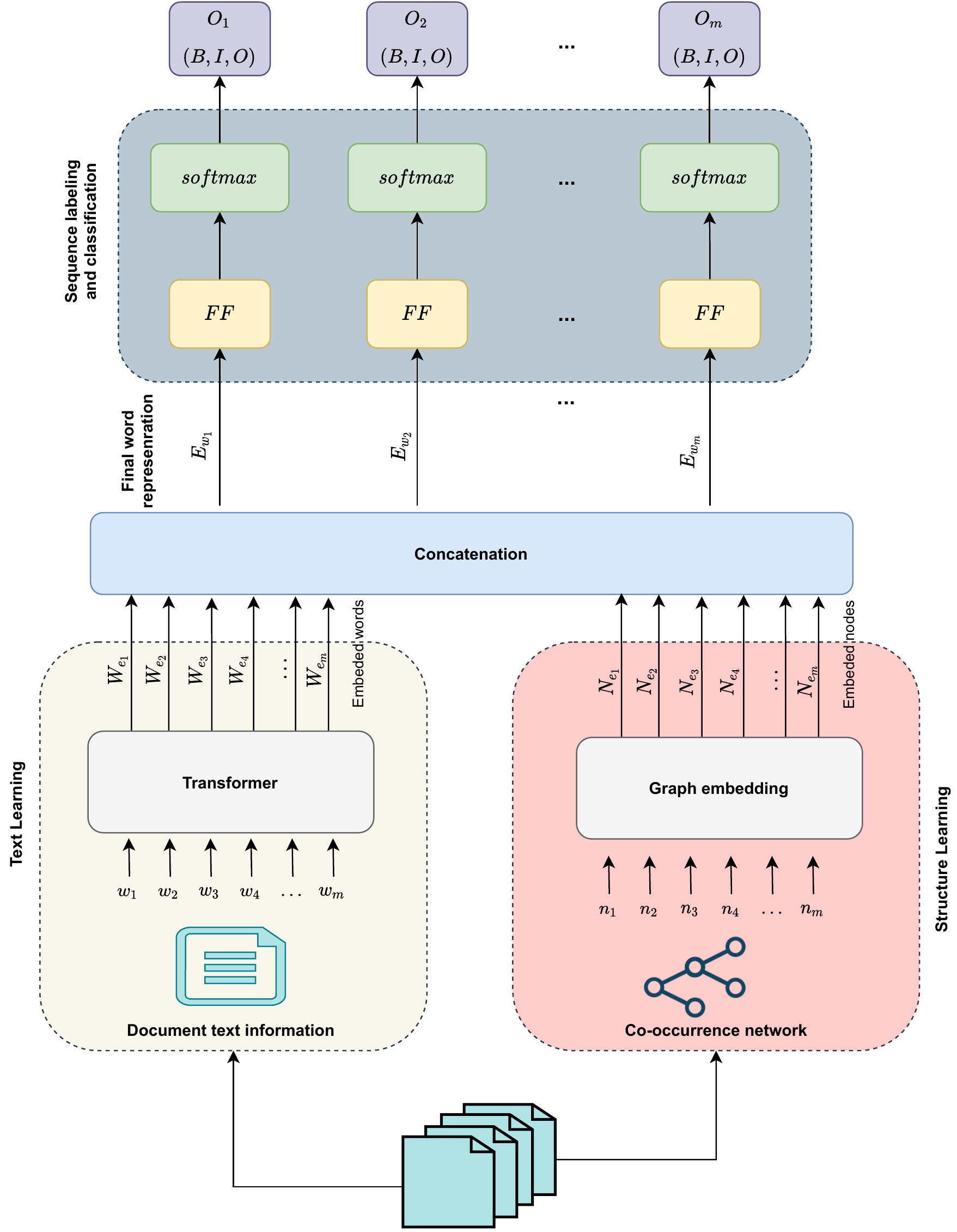}
	\caption{The overall structure of Phraseformer.}
	\label{proposed_method_fig}
\end{figure}

In this section, we propose a framework, called Phraseformer, that extracts key-phrases through learning context representations from text information and node representations in co-occurrence network. The overall processing steps of Phraseformer are explained in Figure \ref{proposed_method_fig}. The flow of framework  is composed of four main
steps, as follows:

\textbf{Text learning:} Find a textual representation for all words using BERT Transformer. This part of Phraseformer provides a deeper understanding to evaluate semantic similarity between words.

\textbf{Structure learning:} Create context of each word from the co-occurrence graph using graph embedding techniques to learn structure representation.

\textbf{Final word representation:} Concatenate the text information and structure information and create a single representation for each word.

\textbf{Sequence labeling and classification:} Formulate key-phrase  extraction  as  a  sequence  labeling  task and label each word based on the BIO tagging scheme through a fully connected layer to classify the word.

\subsection{Text Learning} \label{III-B}
\begin{figure}[ht]
    \centering
    \includegraphics[page=1,width=0.5\linewidth]{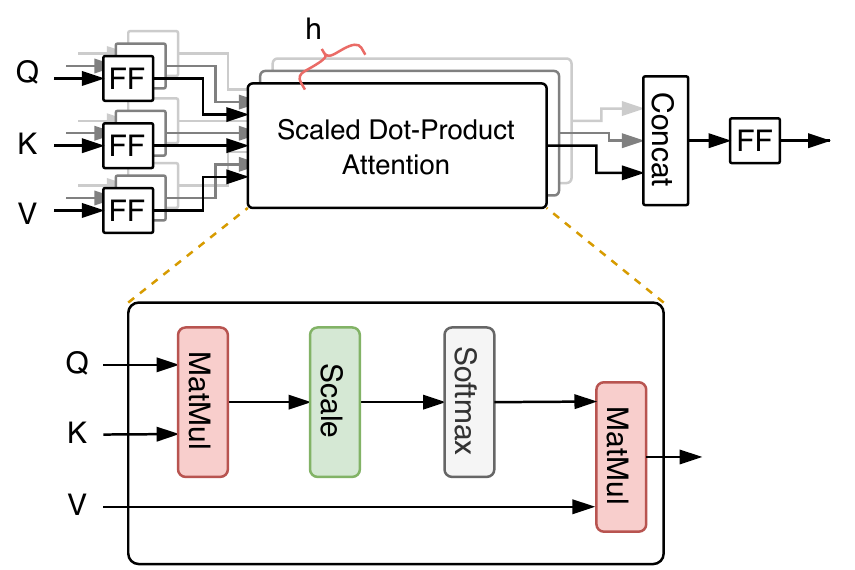}
    \caption{ The Transformer model architecture \cite{asgari2020multimodal}.}
    \label{fig:transformer}
\end{figure}

\begin{figure}[ht]
    \centering
    \includegraphics[page=1,width=0.7\linewidth]{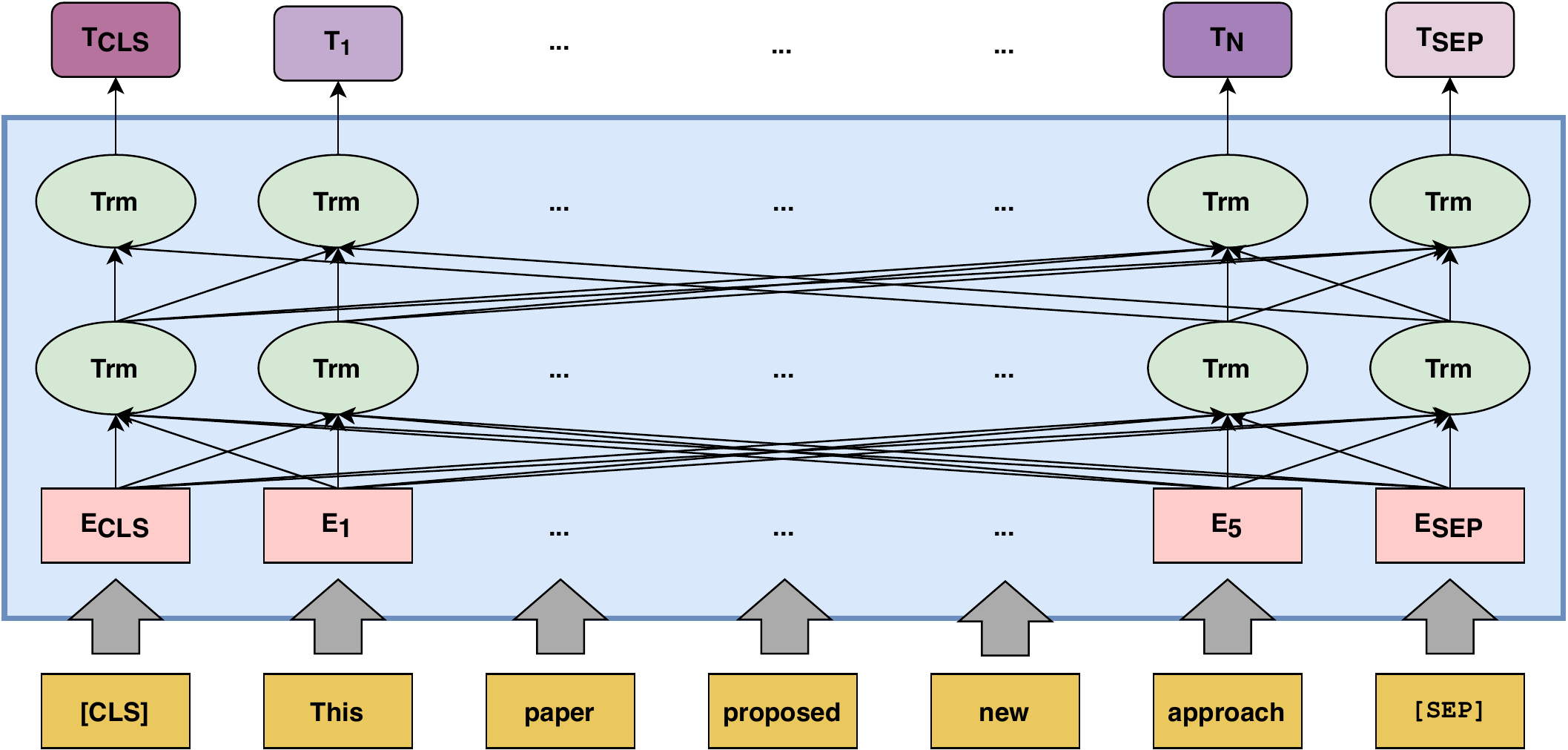}
    \caption{The BERT Transformer architecture \cite{nikzad2020berters}.}
    \label{fig:bert}
\end{figure}

The first step of Phraseformer is to generate the textual vector for every word. With the advent of Transformers, the way of working with text data has truly altered. Transformers eliminate the drawbacks of RNN and CNN architectures. Applying self-attention allows Transformers to be much more parallelization than other architectures \cite{minaee2020deep}. A  Transformer is composed of encoding and decoding components  as shown in Figure \ref{fig:transformer}. An encoding component includes a number of encoder blocks which have two layers: a Multi-Head Attention layer and a Feed Forward Neural Network layer \cite{asgari2020topicbert}. On the other side,  blocks with a Masked Multi-Head Attention layer before the feed forward layer create the decoding component \cite{vaswani2017attention}. Also, both components contain the same number of blocks. There are different models based on Transformer structure such as BERT \cite{devlin2018bert}, OpenGPT \cite{radford2018improving,radford2019language}, XLNet \cite{yang2019xlnet} and ELMo \cite{peters2018deep}. In this study, the technique to learn textual representation is BERT Transformer whose structure is presented in Figure \ref{fig:bert}. One of the important advantages of BERT over models like Word2Vec is that BERT creates the word embedding for each word based on the each sentence or each document the word is in it. That means that BERT is capable of capturing the context of a word in a document. The text information of a document is composed of the documents’ titles,
abstracts. The left block in Figure \ref{proposed_method_fig} shows the process for learning textual word vectors.

\subsection{Structure Learning} \label{III-C}
The second step is to learn the structural vector for every word based on three graph embedding algorithms including ExEm \cite{nikzad2020exem} , Node2vec \cite{Grover2016} and DeepWalk \cite{Perozzi2014} in the co-occurrence network. A co-occurrence network is defined as a graph $G = (V, E)$ where nodes represent words and  each edge $e \in E$ demonstrates the co-occurrence relationship between word $v_i$ and word $v_j$ that appear in a context window. Since the quantity of information that can be acquired from a solitary document is finite, we construct the co-occurrence graph over the all documents instead of a single document, as right block in Figure \ref{proposed_method_fig} shown. It should be noted that before building this graph, we remove the stop words and punctuation mark. Then,  our task is to learn the dimensional latent representations of words from this graph using graph embedding techniques. There are different node representation learning algorithms. According to the results reported by different studies, we select three random walk based graph embedding approaches that present better performance. In DeepWalk, a set of random walks are generated by starting from every node in the graph. While, Node2vec proposes a biased random walk method that is modified version of DeepWalk. This method uses two parameters to control over the search space. Moreover, ExEm is another technique that uses dominating set theory to create random walks. ExEm characterizes the local neighborhoods by starting each path sampled with a dominating node. Also, the global structure information is captured by selecting another dominating node in the random walk. All three approaches inject these random walks into a skip-gram neural network model to learn node representations. 

\subsection{Final Word Representation}
After we obtain each word's vector representation of text information and co-occurrence network structure, the following step is to combine these information into a single representation. We believe that the concatenation of text-based and structure-based information can better discover the potential of words for being keywords. So, we present each word with a solitary vector which is the combination of text and structure vectors. For example for word $w_i$, we show $E_{w_i} = W_{e_i} + N_{e_i}$ where $W_{e_i}$ and $N_{e_i}$ denote the text and structure learning representations for word $w_i$, respectively.

\subsection{Sequence Labeling and Classification}
Sequence labeling is a type of pattern recognition task in NLP domain that categorizes words in a text and assigns a class or label to each word in a given input sequence \cite{he2020survey, akhundov2018sequence}. There are numerous techniques for sequence  labeling  task include Hidden  Markov  Models \cite{kupiec1992robust},  Conditional Random Fields (CRF) \cite{lafferty2001conditional} and deep learning approaches.

In  this  paper, we consider the problem of key-phrase extraction from a document as a sequence labeling task.  Also, we observe the sequence labeling in the form a classification task using the BIO encoding scheme as output labels.
So, our model takes $E_{w_1}, E_{w_2},...,E_{w_m}$ as inputs and assigns each word a label $O_i \in \{ B,I,O \}$ where $B$ shows that $w_i$ is the beginning of a key-phrase, $I$ denotes that $w_i$ is inside a key-phrase, and finally $O$ illustrates that $w_i$ is outside of a key-phrase. It can be conducted from Figure \ref{proposed_method_fig} that
a fully connected structure and a softmax layer are used to make the classification decision.

\section{Experimental Evaluation}\label{experiments}
In this section of our research, we will clarify that which datasets have been used to evaluate Phraseformer. Also, we are going to describe the baseline algorithms to compare our proposed method against them. In the next part, we will present different versions of Phraseformer. Finally, evaluation metrics will be specified.

\subsection{Dataset}
To assess the efficiency of Phraseformer, we used three datasets including Inspec \cite{hulth2003improved}, SE-2010 \cite{kim2010semeval} and SE-2017 \cite{augenstein2017semeval}. We will describe these datasets in the succeeding paragraphs.

\textbf{Inspec} includes abstracts of papers from Computer Science collected between the years 1998 and 2002.

\textbf{SE-2010} contains of full scientific articles that are obtained from the ACM Digital Library. In our experiment, we used the abstract of papers.

\textbf{SE-2017} consists of paragraphs selected from 500 ScienceDirect journal papers from Computer Science, Material Sciences and Physics domains.

It should be noted that because of formulating the key-phrase extraction as a sequence labeling task, we take into consideration key-phrases that appear in the abstracts of articles in three datasets. Table \ref{dataset_detail}  shows  the  statistics\footnote{We obtained this information from \url{https://github.com/LIAAD/KeywordExtractor-Datasets}}.  of  the  above three  datasets.

\begin{table}[ht]
\caption{Dataset statistics.}
\label{dataset_detail}
\begin{tabular}{l|l|l|l|l|l}
\hline
\textbf{Dataset} & \textbf{Type of doc} & \textbf{\#Doc} & \textbf{\#Gold Keys (per doc)} & \textbf{\#Tokens per doc} & \textbf{Absent Gold Key} \\\hline
Inspec           & Abstract             & 2000           & 29230 (14.62)                  & 128.20                    & 37.7\%         \\
SemEval2010      & Paper                & 243            & 4002 (16.47)                   & 8332.34                   & 11.3\%         \\
SemEval2017      & Paragraph            & 493            & 8969 (18.19)                   & 178.22                   & 0.0\%      
\\ \hline
\end{tabular}
\end{table}

\subsection{Baseline models}
In this study, we compared Phraseformer method with four graph-based approaches and three textual methods. In the following paragraphs, we will explain these techniques in more detail.

\textbf{TextRank} \cite{mihalcea2004textrank} is the simplest graph-based method that is based on PageRank algorithm.

\textbf{DeepWalk} \cite{Perozzi2014} is a graph embedding technique based on random walks. This method represents each node as a low-dimensional vector.

\textbf{Node2vec} \cite{Grover2016} is modified version of DeepWalk that uses a biased random walks to convert nodes into vectors.

\textbf{ExEm} \cite{nikzad2020exem}  is a random walk based approach that uses dominating set theory to generate random walks.  ExEm$_{w2v}$ and ExEm$_{ft}$ are two different version of ExEm.

\textbf{Word2Vec} \cite{mikolov2013efficient} learns the vector representations of words. This method passes the document through two feed-forward layers to created vectors \cite{minaee2020deep}.

\textbf{BiLSTM-CRF} \cite{sahrawat2019keyphrase} is a textual method that considers the keyphrase extraction as a sequence labeling task using a BiLSTM-CRF  architecture.

\textbf{BERT} \cite{devlin2018bert} is a textual approach that uses the transformer structure to obtain the document representation.

\subsection{Method Variations}
We present four variations of Phraseformer that employ different graph embedding techniques for structure learning. Phraseformer(BERT, DeepWalk), Phraseformer(BERT, Node2vec), Phraseformer(BERT, ExEm$_{w2v}$) and Phraseformer(BERT, ExEm$_{ft}$) are different models of Phraseformer that use DeepWalk, Node2vec, ExEm$_{w2v}$ and ExEm$_{ft}$ approaches to obtain word representation from the co-occurrence graph, respectively.

\subsection{Metrics}
To measure the experimental effect, we used F1-score that is formulated as follow:

\begin{equation}\label{eq:dcg}
	F1-{score}=2\times \frac{\frac{Y \cap Y'}{Y'}\times\frac{Y \cap Y'}{Y}}{\frac{Y \cap Y'}{Y'}+\frac{Y \cap Y'}{Y}}
\end{equation}

here $Y'$ and $Y$ are the predicted keywords and the real keywords, respectively.

\section{Experimental Results}\label{evaluation_re}
This section resents  the  experimental results. Firstly, we compare Phraseformer against of baselines. Then, we test different classifiers for the classification part and report consequences.

\subsection{Baseline comparisons}
we compare our multimodal model against textual and graph-based methods. Table \ref{baselines_tb} presents the results. As expected, our model significantly outperforms all the methods. From the results we have the following observations:
\begin{enumerate*}[label={\roman*)},font={\bfseries}]
    \item It can be concluded that Phraseformer approaches gain the highest F1-scores. Phraseformer(BERT, ExEm$_{ft}$) strengthens the performance by $6.4 \%$, $19.94 \%$ and $13.70 \%$ compared with BERT over Inspec,   SE-2010  and SE-2017 datasets, respectively.  Moreover, Phraseformer(BERT, ExEm$_{ft}$) outperforms ExEm  by $112.6 \%$, $290.4 \%$ and $130.4 \%$ over the mentioned datasets. These high scores prove that our hypothesis about using multimodal learning, sequence labeling and classification is true.
	\item It is evident that two graph-based and textual methods indicate poor performances.
	\item It is obvious that textual methods have satisfactory outcomes compared with graph-based models.
    \item Moreover, BERT shows better results than other textual methods. Because BERT embeds words into vectors by considering the meaning of the words in the sentence.
    \item Additionally, the results of graph-based methods demonstrate that the graph embedding techniques gain the highest F1 scores than TextRank. Also, various versions of ExEm are more successful than other graph embedding approaches. The reason is that ExEm obeys the homophily and structural role equivalences in learning node representations with help of dominating nodes.
\end{enumerate*}

\begin{table}[H]
\centering
	\caption{Comparison with baseline methods (F1-score).}
	\label{baselines_tb}
	\begin{tabular}{p{1.5cm}|p{4.4cm}|p{1.5cm}|p{1.5cm}|p{1.5cm}}
\hline
\multirow{2}{*}{Category} &\multirow{2}{*}{Model} & \multicolumn{3}{c}{Dataset}\\
\cline{3-5}
&&Inspec  &   SE-2010   & SE-2017 \\
\hline
\multirow{5}{*}{\rotatebox[origin=c]{90}{\parbox[c]{2cm}{\centering Graph-based method}}} &TextRank \cite{mihalcea2004textrank}& 0.1780 & 0.1990 & 0.2090 \\ 
&DeepWalk \cite{Perozzi2014}& 0.3190 & 0.1102 & 0.2887\\ 
&Node2vec \cite{Grover2016}&  0.3138 & 0.1098 & 0.2863\\ 
&ExEm$_{w2v}$ \cite{nikzad2020exem}& 0.3273 & 0.1233 & 0.2911\\ 
&ExEm$_{ft}$ \cite{nikzad2020exem}& 0.3286 & 0.1246 & 0.2913 \\ \hline
\multirow{3}{*}{\rotatebox[origin=c]{90}{\parbox[c]{2cm}{\centering Text-based method}}} &Word2Vec \cite{mikolov2013efficient}& 0.4730 & 0.2080 & 0.2920 \\
& BiLSTM-CRF \cite{sahrawat2019keyphrase} & 0.5930 & 0.3570 & 0.5210 \\
& BERT \cite{devlin2018bert} & 0.6564 & 0.4056 & 0.5904\\ \hline
%&M-GCKE \cite{wang2019detecting} & 0.3056 & 0.091 & 0.168 \\ 
\multirow{4}{*}{\rotatebox[origin=c]{90}{\parbox[c]{2cm}{Hybrid method}}} &Phraseformer(BERT, DeepWalk) & 0.6844 & 0.4722 & 0.6570\\ 
&Phraseformer(BERT, Node2vec)& 0.6868 & 0.4746 & 0.6594 \\ 
&Phraseformer(BERT, ExEm$_{w2v}$)& 0.6970 & 0.4848 & 0.6696\\ 
&Phraseformer(BERT, ExEm$_{ft}$)& \textbf{0.6987} & \textbf{0.4865} & \textbf{0.6713} \\ \hline
\end{tabular}
\end{table}

\subsection{Classifier}
In this part of our experiment we aim to investigate which classifier is best suited for sequence labelling and classification tasks to find key-phrases. Table \ref{classifier_re_tb} illustrates the performance of different classifiers on Phraseformer method over Inspec dataset. From the results, it can be seen that Random Forest classifier is obtained the best performance among all classifiers, as shown by the underlined result. On the other hand, the bold results in this table present which models of Phraseformer can gain the highest values on each classifier. It is obvious that the combination of BERT and ExEm is more  meaningful and can better represent the semantic of words. Hence, Phraseformer can accurately extract keywords. 

\begin{table}[H]
\centering
	\caption{Classifiers comparison on Inspec dataset (F1-score)}
	\label{classifier_re_tb}
	\begin{tabular}{p{4.4cm}|p{3cm}|p{1.5cm}|p{3cm}|p{3cm}}
\hline
\multirow{2}{*}{Model} & \multicolumn{4}{c}{Classifier}\\
\cline{2-5}
&Random Forest  &  SVM   & Logistic regression & Fully connected \\
\hline
Phraseformer(BERT, DeepWalk)& 0.7024 & 0.6564& 0.6714& 0.6844\\ 
Phraseformer(BERT, Node2vec)& 0.7048& 0.6588 & 0.6738& 0.6868\\ 
Phraseformer(BERT, ExEm$_{w2v}$)&0.7150& \textbf{0.6707}& \textbf{0.6857}& 0.6970 \\ 
Phraseformer(BERT, ExEm$_{ft}$)& \underline{\textbf{0.7167}}& 0.6690& 0.6840& \textbf{0.6987}\\ \hline

\end{tabular}
\end{table}

\section{Conclusion} \label{sec:conclusion}
In this paper, a multimodal approach, Phraseformer, to extract key-phrases from documents is proposed. In this approach, two modalities originate from the co-occurrence graph and the content of documents. Moreover, we formulate the key-phrase  extraction as a sequence  labeling task solved using a classification model. To represent information of the modalities as vectors, we use BERT Transformer and graph embedding techniques. Utilization of graph based information with aid of textual semantics provide a more insightful representation of the keywords that yield into a more robust keyphrase extraction method. Finally, to validate the effectiveness of the proposed Phraseformer approach, we conduct experiments  on  three  datasets,  and  the results demonstrate how Phraseformer significantly outperforms single-modality methods.

\section*{Acknowledgment}
This project is supported by a research grant of the University of Tabriz (Number S/806).

\section*{Declarations}
\subsection*{Funding}
This project is supported by a research grant of the University of Tabriz (Number S/806).
\subsection*{Conflicts of interests}
The authors have no conflicts of interest to declare that are relevant to the content of this article.
\subsection*{Ethical approval}
This article does not contain any studies with human participants or animals performed by any of the authors.

\bibliography{mybibfile}

\end{document}